\documentclass[sigconf]{acmart}
\AtBeginDocument{%
  \providecommand\BibTeX{{%
    \normalfont B\kern-0.5em{\scshape i\kern-0.25em b}\kern-0.8em\TeX}}}

\usepackage{graphicx}
\usepackage{caption}
\usepackage{subcaption}

\usepackage{multirow} 
\usepackage{xspace}

\usepackage{algorithm}
\usepackage{algpseudocode}
\usepackage{url}
\usepackage{soul,xcolor}

\setcopyright{acmcopyright}
\copyrightyear{2022}
\acmYear{2022}
\acmDOI{XXXXXXX.XXXXXXX}

\acmConference[ICCAD'2022]{ACM Conference}{30 October - 3 November 2022}{San Diego, California, USA}
\acmPrice{15.00}
\acmISBN{978-1-4503-XXXX-X/18/06}

\title{EVE: Environmental Adaptive Neural Network Models for Low-power Energy Harvesting System}

\newcommand{\tsc}[1]{\textsuperscript{#1}}

\author{Sahidul Islam\tsc{1}\tsc{*}, Shanglin Zhou\tsc{2}\tsc{*}, Ran Ran\tsc{3}, Yu-Fang Jin\tsc{4}, Wujie Wen\tsc{3}, Caiwen Ding\tsc{2}, Mimi Xie\tsc{1}}
\affiliation{
  \institution{\tsc{1}Department of Computer Science, The University of Texas at San Antonio}
  \institution{\tsc{2}Department of Computer Science, University of Connecticut}
  \institution{\tsc{3}Department of Electrical and Computer Engineering, Lehigh University}
  \institution{\tsc{4}Department of Electrical and Computer Engineering, The University of Texas at San Antonio}
  \country{}
}
\email{{sahidul.islam, yufang.jin, mimi.xie}@utsa.edu; {shanglin.zhou, caiwen.ding}@uconn.edu; {rar418, wuw219}@lehigh.edu}

\begin{document}
\begin{abstract}
IoT devices are increasingly being implemented with neural network models to enable smart applications. Energy harvesting (EH) technology that harvests energy from ambient environment is a promising alternative to batteries for powering those devices due to the low maintenance cost and wide availability of the energy sources. However, the power provided by the energy harvester is low and has an intrinsic drawback of instability since it varies with the ambient environment.  
This paper proposes EVE, an automated machine learning (autoML) co-exploration framework to search for desired multi-models with shared weights for energy harvesting IoT devices. Those shared models incur significantly reduced memory footprint with different levels of model sparsity, latency, and accuracy to adapt to the environmental changes. 
An efficient on-device implementation architecture is further developed to efficiently execute each model on device. A run-time model extraction algorithm is proposed that retrieves individual model with negligible overhead when a specific model mode is triggered. 
Experimental results show that the neural networks models generated by EVE is on average 2.5X times faster than the baseline models without pruning and shared weights.  
\end{abstract}


\maketitle

{
\renewcommand{\thefootnote}{\fnsymbol{footnote}}
\footnotetext[1]{Both authors contributed equally to the paper}
}
\section{Introduction}
In the wake of democratization of artificial intelligence, there are increasing demands on executing deep neural network (DNN) models on sensing devices for better accuracy and more efficient prediction~\cite{garvey2018framework,li2019edge} for various IoT applications~\cite{he2016deep,devlin2018bert,zhou2021end}. 
However, when DNN models come to on-board, there is a grand challenge to accommodate the giant models to tiny IoT devices with limited memory and computing resources \cite{memwarednn,islamenabling,peng2021binary,mendis2021intermittent,kang2020everything,kang2022more}.
Particularly, first, embedded IoT devices have limited computational units and low CPU frequency (e.g., 1-16MHZ). Since DNNs are computationally expensive, DNN algorithm takes long on-board execution time. Second, embedded IoT devices are equipped with small memory (e.g., hundreds of KBs) which can not even afford tiny DNN models (e.g., Tens of MBs). Third, these battery-powered devices naturally have a limited standby time.

Energy harvesting technology that harvests energy from ambient environment is a promising alternative due to the low maintenance cost and wide availability of the energy sources.  However, the power provided by the energy harvester is low and has an intrinsic drawback of instability since it varies with the ambient environment. For example, solar cells can generate power of different densities ranging from 0 to 15$mW/cm^{2}$ depending on the varying light intensity. With an unstable low power supply, the DNN execution will be interrupted frequently, resulting in significantly increased execution time. As a result, a deployed model that has 1 second real-time performance when there is high power intensity may execute for 10s when the power intensity is low, resulting in dramatically degraded quality-of-service (QoS).

This paper proposes EVE, a novel pattern pruning based framework that generates and implements multiple hardware-friendly models with different sparsity but shared weights to adapt to the varying environment of the energy harvesting devices. 
Multiple shared weights models (\textit{SWM}) can successfully fit within the on-chip memory budget \textcolor{black}{which occupies much less memory space than multiple individual models without shared-weights}. For example, if we consider three energy levels as High, Medium, and Low, then \textit{SWM} should consist of three shared-weight models and holds almost the size of only one model. Since our final objective is performing environment adaptive inference, after we deploy the \textit{SWM}, depending on the availability of energy, a particular fit model inside the \textit{SWM} needs \textit{extraction}. Hence, we further introduce \textit{bit-matrix}, which expresses the applied pattern with small overhead. With the help of the \textit{bit-matrix}, a novel model extraction algorithm is proposed that can successfully reconstruct the required individual model from the \textit{SWM} during run-time.


To the best of our knowledge, this is the first hardware/software co-design attempt to adaptively configure DNN models on resource limited energy harvesting devices. The major contributions are summarized as follows:
\begin{itemize}
\item {To satisfy real-time  constraint under the varying harvesting power, we propose a shared weight training method by generating three models, which maximally mitigates the writing and reading energy consumption.}

\item {We propose a AutoML-based framework to search for multiple compressed models with shared weights and best possible accuracies.}
A hardware performance predictor is developed to estimate the inference latency of models when applying different pruning patterns with different sparsity.

\item {We propose a run-time model extraction algorithm to reconstruct a particular model without any information loss or any significant overhead. Our algorithm can instinctively select the candidate model for extraction, based on the energy condition of the environment.}
\end{itemize}

\textcolor{black}{The remainder of this paper is organized as follows: Section II provides the motivation and related work; Section III provides the framework overview and shared-weight training process which is conducted by AutoML search algorithm; Section IV presents the on-device implementation of shared-weight model and adaptive inference; Section V contributes the experimental evaluation, and Section VI concludes this work.}

\section{Motivation and Related Work}
\subsection{Motivation}

\begin{figure}[h]
\includegraphics[scale=0.35]{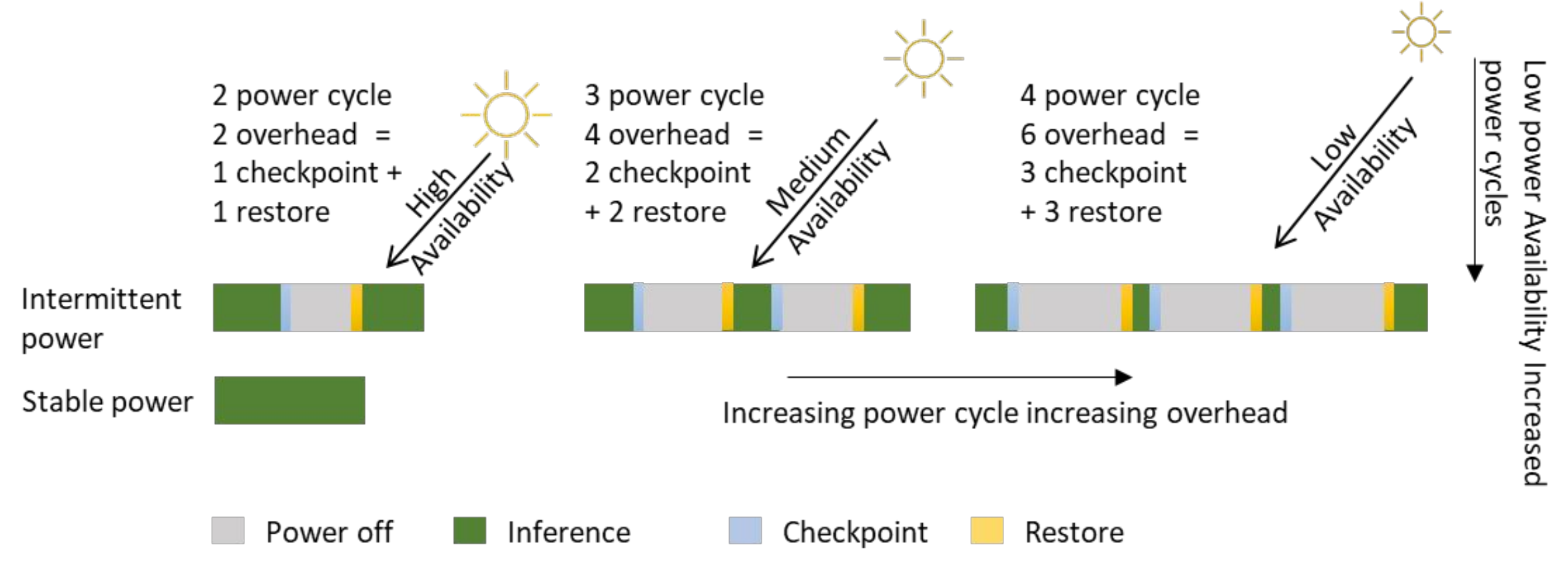}
\vspace{-10pt}
\caption{Illustration of required power cycles under three energy harvesting scenarios.}
\label{Fig:Moti2}
\end{figure}

\textbf{\textit{Why \textcolor{black}{multiple adaptive models} matter?}} Energy harvesting (EH) that harvests energy from ambient environment has an intrinsic drawback: the harvesting power varies with the ambient environment. As a result, the same computation task may take different time to complete under different harvesting environment. Figure \ref{Fig:Moti2} illustrates three different DNN inference scenarios when the energy availability level is high, medium, and low respectively. During a power cycle, an energy harvesting device has 4 states including DNN inference, checkpoint, power off, and restore. Compared with high available energy environment, it takes more power cycles for the DNN inference to complete when there is low available energy due to frequent power failures, high checkpoint/restore overhead, and slow charging rate, resulting in degraded QoS and even failure in meeting the time requirement. 
To meet the time requirement, a promising solution is switching to a low latency model with slightly lower prediction accuracy when there is low available energy. 
If there are multiple models to choose from, then we can dynamically switch the DNN models with different latency and energy consumption according to the energy levels to accommodate the varying
environment during run-time.

To meet the time requirement, the accuracy can be slightly compromised to reduce the computation latency. As we know that the lowest latency and highest accuracy is hard to achieve simultaneously, a trade-off point is needed to achieve the best possible accuracy while satisfying the QoS for a specific environment. 

\vspace{6pt}
\noindent\textbf{\textit{Memory is a major concern for multiple models.}} 
Multiple models need to be deployed into the same resource limited device to adapt to the environment. This is impractical for most of the energy harvesting devices because of scarce resource where a single model needs rigorous compression to deploy. 
\textcolor{black}{Given aforementioned conditions, the question of how to afford multiple model at a time in a single EH devices arises. In this research, we explore pattern-based pruning for designing $SWM$ architecture. 
This paper attempts to achieve maximum commonality of multiple pruned DNN models with different sparsities and the best possible accuracy. 
}



\subsection{Related work}
To enable DNN on small IoT devices, many DNN implementation frameworks have been proposed based on model pruning and neural architecture search techniques. PatDNN~\cite{niu2020patdnn} proposed a real-time DNN execution on mobile devices, which applied architecture aware optimizations to fine-grain pruning patterns for the neural network. SMOF~\cite{liu2021smof} put more effort into reducing kernel size and the number of filter channels to overcome fixed-size width constraints in SIMD units. For AutoML framework on edge devices, \cite{song2021dancing} combined hardware and software reconfiguration through reinforcement learning to explore a hybrid structured pruning for Transformer. Similarly, \cite{qi2021accelerating} designed an algorithm-hardware closed-loop framework to efficiently find the best device to deploy the given transformer model. NeuroZERO introduced a co-processor architecture consisting of the main microcontroller that executes scaled-down versions of a (DNN) inference task~\cite{neuroZero}. TF-Net pipeline efficiently deploys sub-byte CNNs on microcontrollers~\cite{tfNet}. A software/hardware co-design technique that builds an energy-efficient low bit trainable system is proposed in~\cite{lowbit}. All the above works can enable DNN implementation on battery powered devices.

Due to the advantages of energy harvesting technologies, implementation of DNN models on intermittently powered devices have been proposed. SONIC is an intermittence-aware software system with specialized support for DNN inference~\cite{intelBeyondEdge}. ACE is the accelerator based fast intermittent DNN inference on EH device \cite{islamenabling}. Intermittent-Aware Neural Architecture Search for task based inference is proposed in \cite{mendis2021intermittent}. Hardware Accelerated Intermittent inference is conducted in \cite{kang2020everything}. \textcolor{black}{Model augmentation technique is proposed to adapt DNNs to intermittent devices \cite{kang2022more}.} 


Different from the existing works, this paper considers environmental changes of harvesting power for the first time and configures multiple models at a time on energy harvesting (EH) devices. 
\section{Hardware-aware shared-weight models search framework}

\begin{figure*}[ht]
\center
\includegraphics[width=1.9\columnwidth]{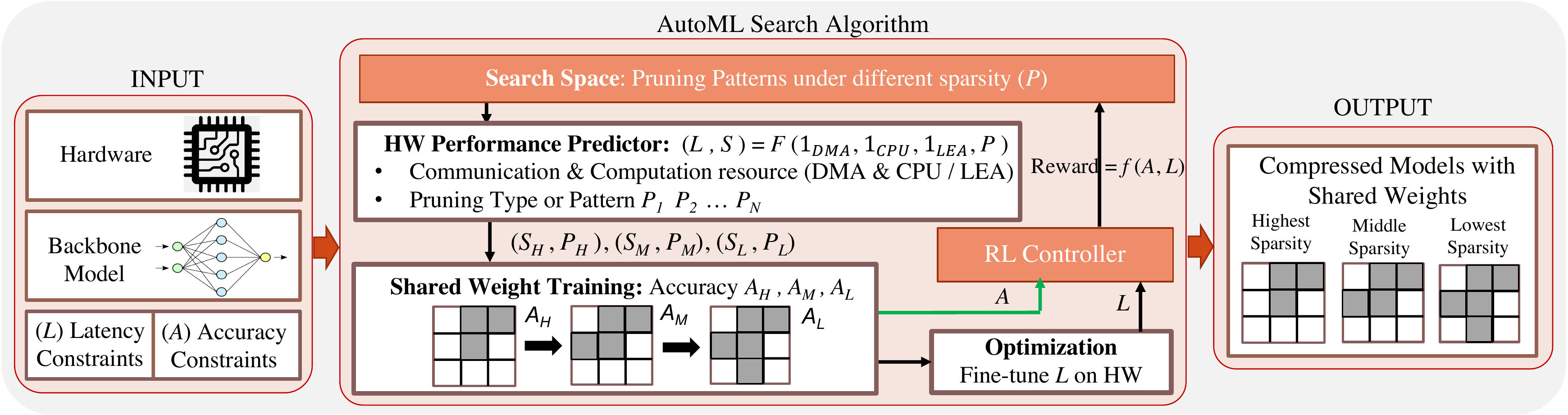}
\caption{Overview of the proposed framework.}
\label{Fig:ovrw}
\end{figure*}


Fig.~\ref{Fig:ovrw} shows the overview of the proposed co-design framework, to satisfy real-time constraints under varying harvesting power. 
Given the hardware resources and the latency ($L$) and accuracy ($A$) constraints as inputs, the proposed AutoML search algorithm uses an RNN-based reinforcement learning (RL) controller to guide searching for the best set of shared-weight compressed models starting with a backbone model. 

In the framework, the AutoML first generates $N$ pruning pattern samples ($P$) under different sparsity. We then abstract a hardware performance predictor to estimate the inference latency of the $N$ models that applied the sampled pruning patterns on the given hardware. Then a hierarchical shared weight training architecture is designed to train these $N$ compressed models. These $N$ models are under different compression ratios, which can be deployed to the different energy levels of the low-power energy devices. 
The $N$ models with shared weights are then fine-tuned for resource allocation and are optimized for the latency under the given hardware constraints and restrictions. Different rewards are given under different constraints satisfaction situations. The controller is then updated based on the feedback (reward) and predicts a better pruning pattern set. 
In the following subsections, we will introduce each component in Fig.~\ref{Fig:ovrw}.

\subsection{Hardware Performance Predictor}

\begin{figure}
     \begin{subfigure}[b]{0.32\columnwidth}
        \hspace{-5pt}
        \vspace{-5pt}
         \includegraphics[width=3cm,height=3cm]{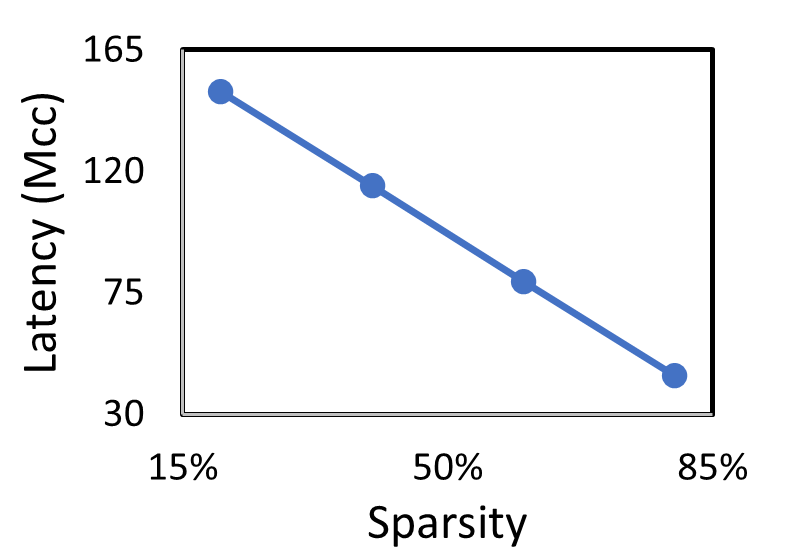}
         \caption{CPU}
         \label{fig:graph1}
     \end{subfigure}
     \hfill
     \begin{subfigure}[b]{0.32\columnwidth}
         \hspace{-9pt}
         \vspace{-5pt}
         \includegraphics[width=3cm,height=3cm]{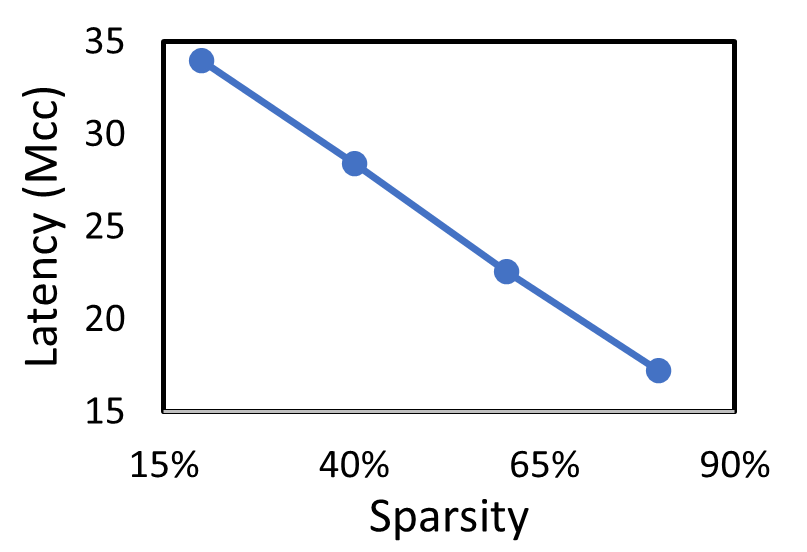}
         \caption{LEA irregular}
         \label{fig:graph2}
     \end{subfigure}
     \hfill
     \begin{subfigure}[b]{0.32\columnwidth}
         \hspace{-9pt}
         \vspace{-5pt}
         \includegraphics[width=3cm,height=3cm]{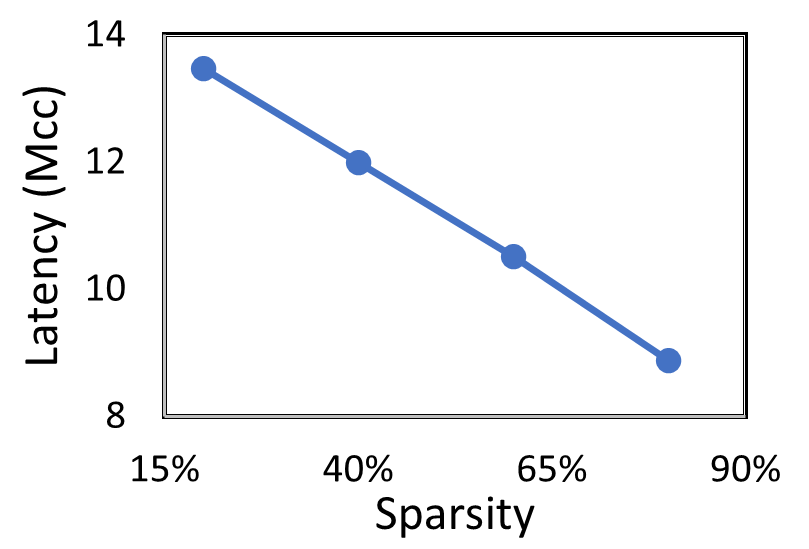}
         \caption{LEA regular}
         \label{fig:graph3}
     \end{subfigure}
     \vspace{-8pt}
        \caption{Latency predictor graph.}
        \label{fig:graphs}
\end{figure}

We develop a hardware performance predictor to estimate the inference latency of models when applying different pruning patterns with different sparsity. For each single DNN layer, we derive the sparsity vs latency curve by implementing different combinations of patterns that vary with sparsity and the hardware computation resources. Figure \ref{fig:graphs} shows an example curve of the relation between latency and sparsity when different computing units are utilized for the DNN inference.

Based on the obtained curves, we produce latency-profiler functions for regular pattern pruning and irregular pattern pruning respectively. Execution with low energy accelerator (LEA) is faster when the applied patterns are regular. However, in CPU based inference for MSP430 devices, execution speed does not depend on pattern regularity since CPU does not perform bulk computation as LEA. Regular patterns lower the data movement cost resulting in reduced latency compared with irregular patterns.

The reason we develop the latency-profiler function is that there are too many patterns with different sparsity and shape combinations. Implementing all these combinations is extremely time expensive and pointless since a simple linear regression can closely predict the latency. Therefore, our performance predictor takes pruning patterns as input, analyzes layers in the given backbone model, and uses the latency-profiler functions to estimate the latency of the entire compressed model.



\subsection{Shared Weight Training}
\label{seq}



\begin{figure}[h]
\centering
\includegraphics[width=0.48\textwidth]{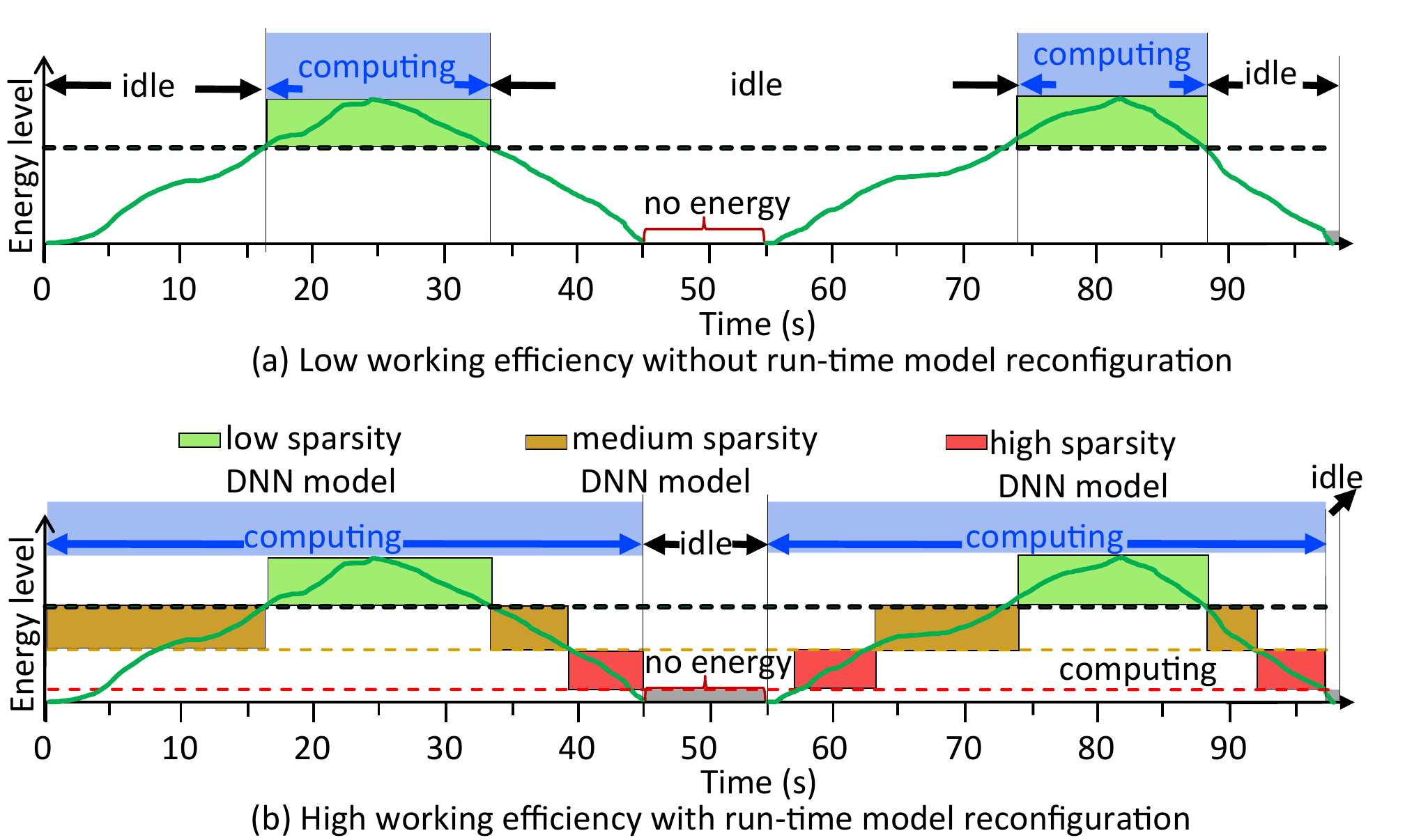}
\vspace{-8pt}
\caption{Run-time dynamic model switch under intermittent power trace in energy harvesting devices.}
\label{Fig:runtime}
\end{figure}

We can dynamically switch the DNN models (\textbf{\textit{with different sparsities but shared weights}}) according to different energy levels. Different sparsities (e.g., high, medium, low) can always enable energy devices to operate even when energy level is low, while the shared weights can minimize the writing overhead when switching to a different model. Overall, we prolong the "working efficiency" of the self-powered devices on the intermediate power trace, shown in Fig.~\ref{Fig:runtime}.

\begin{figure}[h]
\center
\includegraphics[width=0.8\columnwidth]{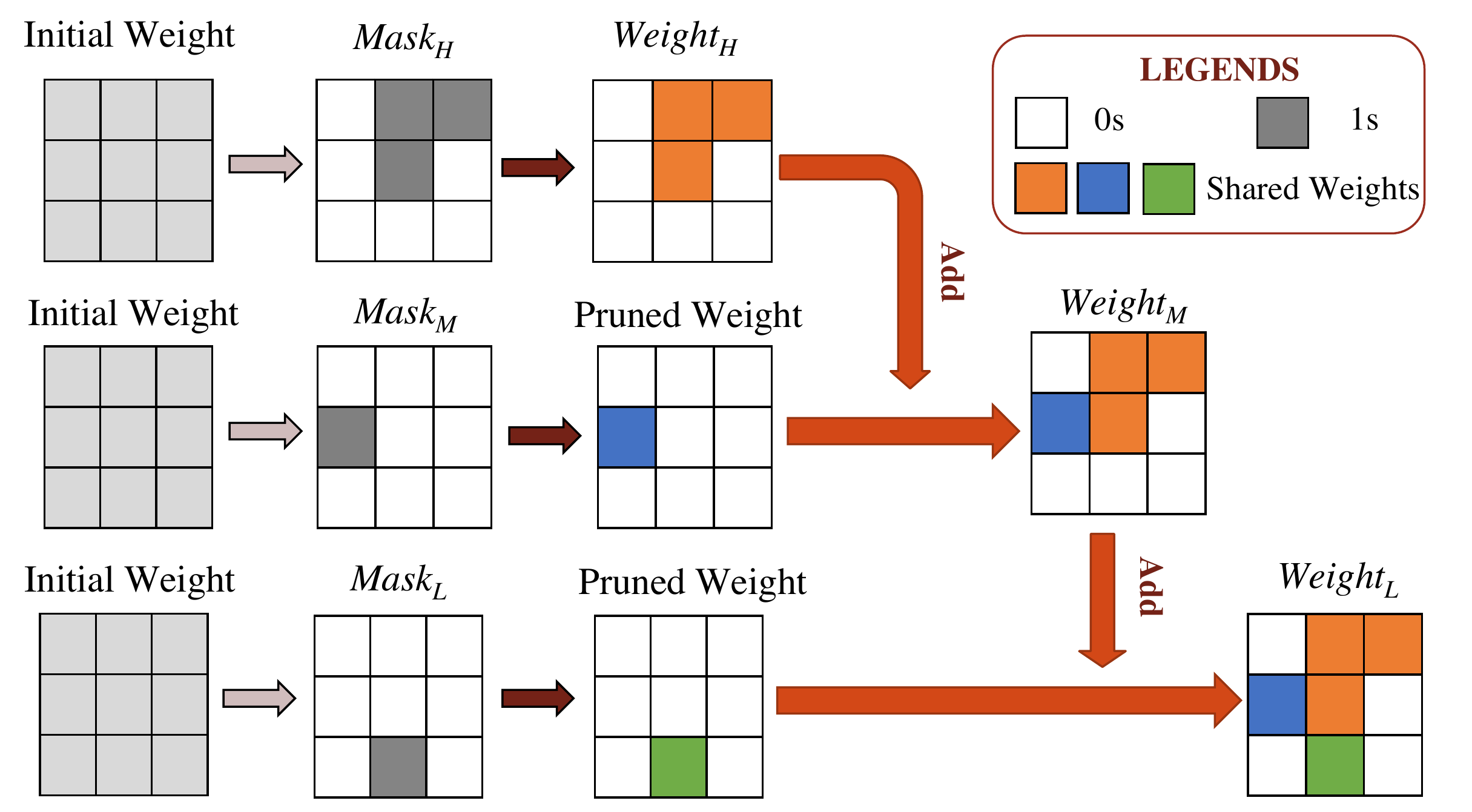}
\caption{Shared weights training workflow.}
\label{Seq}
\end{figure}

\noindent\textbf{Shared weights to mitigate writing overhead.}
An intuitive solution is to directly train several DNN models with different sparsities (e.g., high, medium, low). However, when we switch different DNN models according to different energy level, there will be a large amount overhead due to the weight re-written on memory.
To reduce the overhead, we will 
generate different models
with shared weights, as illustrated in Fig.~\ref{Seq}. The white cells in the figure represent 0s or pruned weights. The grey cells represent unpruned weights. Orange, green and blue cells are remaining weights after pruning. The cells in the same green or red color stand for shared weights. 
When switching between models, we only need to update partial weights, and thus 
we reduce the writing overhead caused by dynamic model switch under the intermittent power trace.

In our shared-weights training (for demonstration, we use three models), the weights of the pruned layer are:
\begin{itemize}
    \item 
       $High Sparsity: W_{h} = W_{Initial} \times M_{h}$
    \item 
      $Medium Sparsity: W_{m} = W_{Initial} \times M_{hm} + W_{h}$    
    \item 
        $Low Sparsity: W_{l} = W_{Initial} \times M_{hm} + W_{m}$
    
\end{itemize}

In the equations shown above, $W$ denotes the weight matrix, $M$ represents the mask. The $footnote$ stands for the percentage of sparsity. $h$, $m$, $l$ and $hm$ stand for $high$, $medium$, $low$ and $slightly$ $higher$ $than$ $medium$ sparsity, respectively. It will be illustrated further in the Fig.~\ref{Seq}, in which the shared-weights training workflow is illustrated.

The shape of the pattern needs to be selected carefully. Since the central elements of a kernel weighs more importance \cite{niu2020patdnn,zhang2021unified}, to guarantee accuracy, we try to keep the central elements as much as possible. Besides, the number of applied patterns also need to be selected carefully since it must fit into small memory. 





\subsection{AutoML Search Algorithm}
\label{AutoML}
In our design, our search space includes three levels: sparsity level, pruning type level, and pruning pattern level. The search space is very big, and it will contain many combinations if we simultaneously search for $N$ different pruning patterns. In this case, we exploit reinforcement learning to guide the search. We use the RNN from~\cite{zoph2016neural} to implement our RL controller and leverage the idea from~\cite{qi2021accelerating} to design our RL algorithm. 

\begin{figure}[h]
\center
\includegraphics[width=0.9\columnwidth]{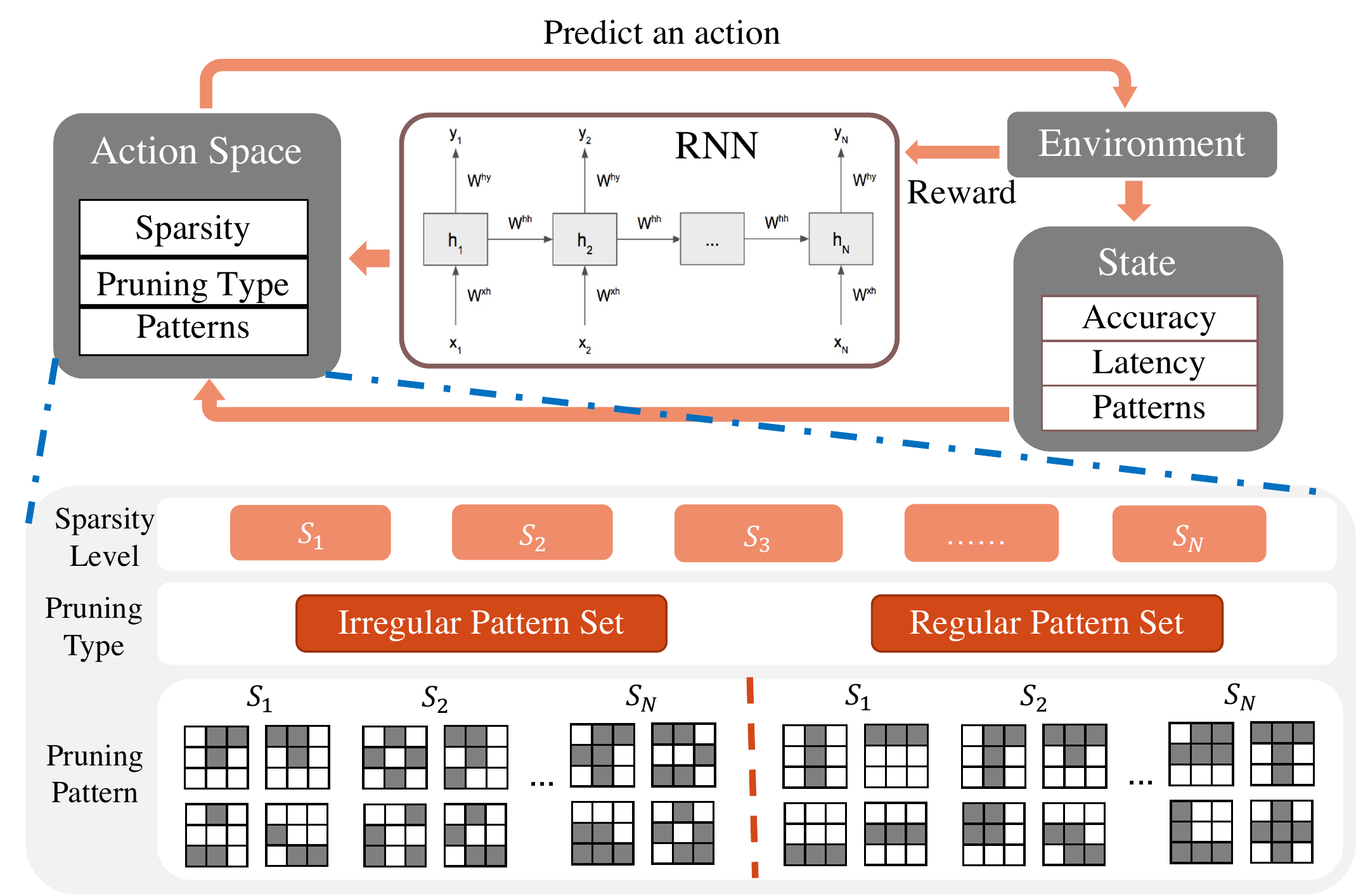}
\caption{RL workflow. Action space includes pruning patterns under different sparsity. The environment is composed of a HW performance predictor and a shared weight training as shown in Fig.~\ref{Fig:ovrw}.}
\label{Fig:RL}
\end{figure}

As shown in Fig.~\ref{Fig:RL}, in each episode, the controller firstly predicts $N$ pruning patterns from the search space for $N$ energy levels. The predicted patterns can be regarded as actions. Then the $N$ pruning patterns are fed to the environment for evaluation. The environment mainly contains three modules. The HW performance predictor module takes the $N$ patterns as input, analysis the sparsity of each pattern, estimates the latency for models when being compressed using the sampled patterns, and verifies whether the latency constraints can be satisfied. The shared weight training module takes the sparsity and patterns as input, trains on the backbone model to obtain accuracy for all the $N$ patterns on a hold-out dataset, and also estimates whether the accuracy constraints can be satisfied. \textit{Reward} will be calculated based on the feedback from the environment. The parameters in the RNN will be updated using the Monte Carlo policy gradient algorithm~\cite{yang2020co} during this period, as follows Eq.~\ref{MC}. 
\begin{equation}
\label{MC}
\nabla J\left(\theta\right)=\frac{1}{K} \sum_{k=1}^{K} \sum_{s=1}^{S} \beta^{S-s} \nabla_{\theta} \log \left(a_{s} \mid a_{(s-1): 1}, \theta\right)\left(R-r\right)
\end{equation}
Here, $\theta$ is the parameters in the RNN, $K$ is the batch size, $S$ is the total number of steps in each episode. $\beta$ is the exponential factor to adjust the reward $R$ at every step, and the baseline $r$ is the average exponential moving of rewards.
With the obtained information, we can formulate the reward function as Eq.~\ref{reward}. 
\begin{equation}
R = \left\{\begin{array}{cc}
A+\frac{L_C-L}{L_C}                & L<L_C, A>A_C \\
-\phi_{P}                         & P \textrm{ not satisfied}    \\
-\phi_{A} \textrm{ or } -\phi_{L} & \textrm{otherwise}
\end{array}\right.
\label{reward}
\end{equation}
Here, $A$ is the lowest accuracy and $L$ is the largest latency of the $N$ models for the $N$ energy levels, $L_C$ and $A_C$ are the given latency and accuracy constraints respectively, $P$ is the predicted pruning pattern, and $\phi$ are predefined numbers that represent constant penalty.
Three cases will occur when calculating the reward $R$. 
(1) if $L<L_C$ and $A>A_C$, it indicates that the performance of the models with the sampled patterns can satisfy the constraints, then we sum up the reward of hardware performance and accuracy; 
(2) if the predicted patterns for the $N$ energy levels are the same or under the same sparsity, then the pattern constraint is not satisfied, so we return negative value $-\phi_{P}$ to the controller; 
(3) in any other cases, which means either latency constraint or accuracy constraint is violated, we also return negative values to the controller. Different penalty are given to guide the search: $-\phi_A$ is set for $L<L_C \;\&\; A<A_C$, $-\phi_L$ is set for $L>L_C \;\&\; A>A_C$.

\section{On-device architecture of shared-weights models}
In this section, we will discuss on-device implementation of the \textit{SWM}.
In pattern based pruning, the pruned information is unnecessary to keep in the weight matrix. Therefore, the condensed weight matrix, consisting only unpruned data can achieve significant memory cutback. Traditionally, sparse matrix and it's condensation requires storing extra information like (row, column, stride, offset etc) \cite{niu2020patdnn}. Storing such information introduces extra storage requirements. In this work, since we are working on pattern-based pruned CNN, we only store the location-index, applied patterns (represented as \textit{bit-matrix}), and shared-weights, which allow us to significantly reduce the memory overhead required in the process of extracting the weight matrix for each individual model.

By following this process, the shared weights found from the AutoML search and shared weight training are further compressed to generate a unified on-device deployable \textit{SWM}.
During inference, we perform the specific model extraction by extracting the condensed weights from the \textit{SWM}. The condensed weights do not require reconstruction, since we extract the corresponding input by applying the pattern associated with the kernel and perform convolution operations as shown in figure \ref{Fig:arch_conv}. To accelerate the computation we used on-device LEA accelerator and DMA data transfer.

\subsection{Inference with \textit{SWM}}
Pattern based pruning opens up the opportunity to deploy multiple pruned models with different sparsities that perform the uniform task.
In Figure \ref{Fig:arch_com_extra}, green model has low sparsity while blue model has high sparsity. Conventionally, the DNN has the accuracy vs speed trade-off and which is why the sparsity differences among those models are the key to speed up the inference or improve the accuracy. For example, a low sparsity model provides higher latency and higher accuracy. \textcolor{black}{On the contrary, high sparsity models decrease the latency and accuracy. This relationship plays a vital role in adaptive inference. }

\subsubsection{Weight Sharing}
To deploy multiple models with different levels of sparsity, our framework ensures two things. First, the autoML search and training process ensures that all the models have shared weight, which is discussed in Section \ref{seq}. Second, our framework attempts to maximize the commonality or shared weight among all the different sparsity models by approaching the idea that high sparsity models are the subset of low sparsity model. For instance, if we design three models (A, B, C) where the applied patterns are (P,Q,R) such that P$>$ Q $>$ R in terms of sparsity. Our aim is to maximize the commonality among P, Q, and R by minimizing the symmetric difference such that $minimize( P \ominus Q\ominus R )$, so that A$>$ B $>$ C in terms of inference speed. \textit{ Thus, in ideal scenarios, large sparsity patterns become the subset of the small sparsity patterns such that $P \subset Q \subset R$}. The ideal scenario is preferable because it allows more compression and low extraction complexity. However, not every kernel of the models can acquire such property since there is an accuracy-pattern trade-off. 

Once such models are achieved after the sequential shared-weight training, we compress and deploy the \textit{SWM} into the device.

\subsubsection{Weight Compression}
\begin{figure}
\includegraphics[width=6cm, height=6cm]{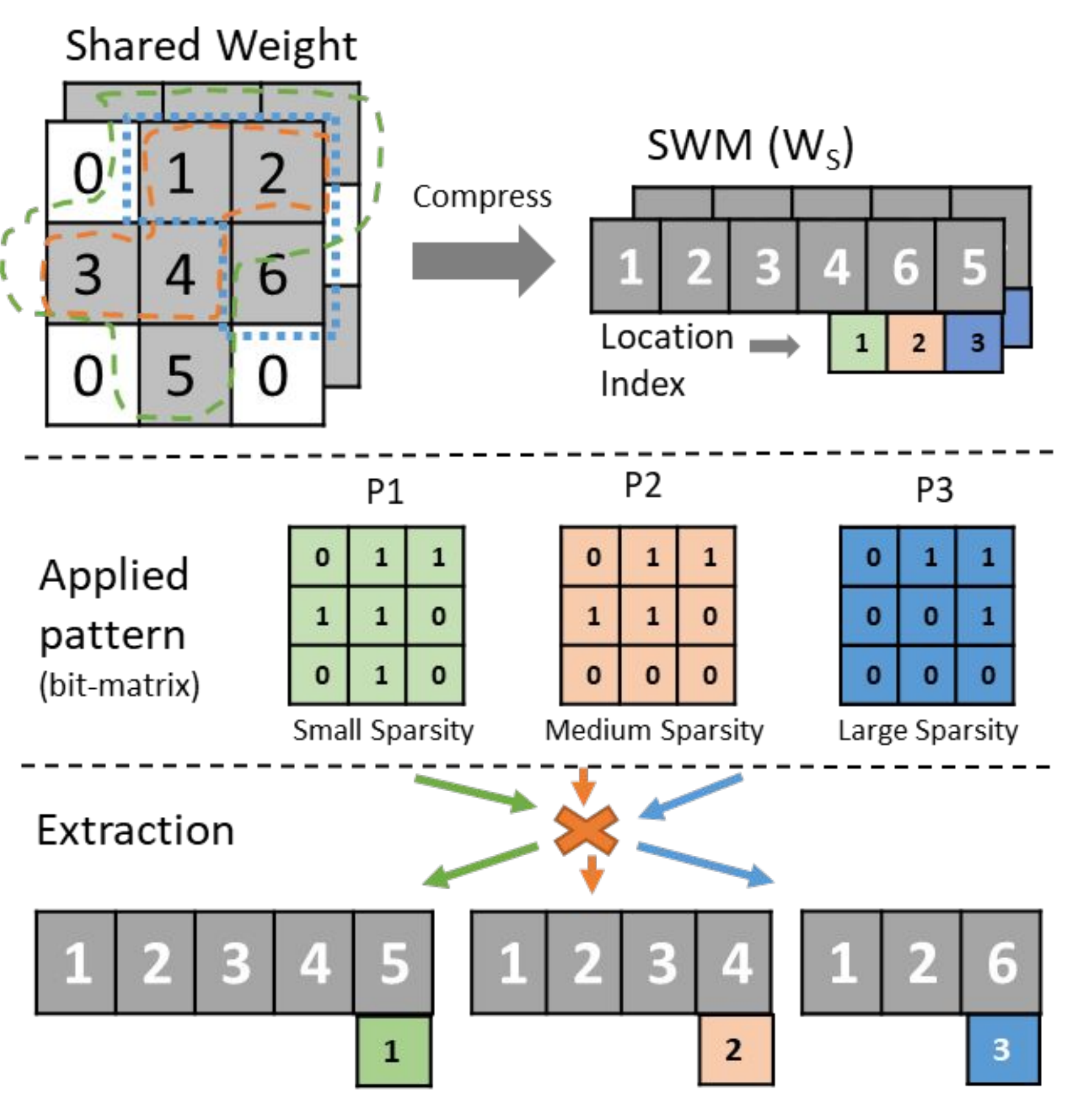}
\vspace{-7pt}
\caption{Shared weights compression and extraction: \textcolor{black}{three shared-weight kernels with different sparsities are compressed to generate 1D \textit{SWM} to be deployed on device; Model specific weights are extracted from the SWM according to the applied patterns.}}
\label{Fig:arch_com_extra}
\end{figure}

We compress all the shared-weight models by eliminating the pruned information from the weight matrix as shown in Figure \ref{Fig:arch_com_extra}. By eliminating pruned information from every model, we get condensed weight matrix. 
We take the unified set of all condensed weights and store them in a 1D array to form compressed weights. 
Besides, the location indices for different patterns are saved to locate the right pattern during runtime. 

The applied patterns are obtained from the pattern search space during autoML search discussed in section \ref{AutoML}. 
Since the number of the applied patterns is limited and fixed for a given model (e.g., 3-6), it takes only few bytes of memory to encode the patterns with  \textit{bit-matrix}. 
A \textit{bit-matrix} represents the applied patterns with 0s and 1s. Here, 0 indicates pruned weight while 1 indicates the kept weight. Each \textit{bit-matrix} is deployed on device with the \textit{SWM}.  With this design, \textit{SWM} occupies much less memory space than multiple models without shared weights.

\textcolor{black}{
\noindent Figure \ref{Fig:arch_com_extra} demonstrates the compression and extraction process with three different sparsity models. \textit{Note that, this process also works when there are more than three models.} Here, green (C), orange (B), and blue (A) represent low (R), medium (Q) and high (P) sparsity model respectively. \textcolor{black}{Since non-ideal scenario is more complex, in this particular example, we demonstrate with a non-ideal scenario where high sparsity model is not the subset of low sparsity models.} Here, Q is a subset of R, but P is not a subset of Q or R since they do not share the weight 6. 
}

\subsubsection{Weight extraction}
\begin{figure}
\includegraphics[scale=0.37]{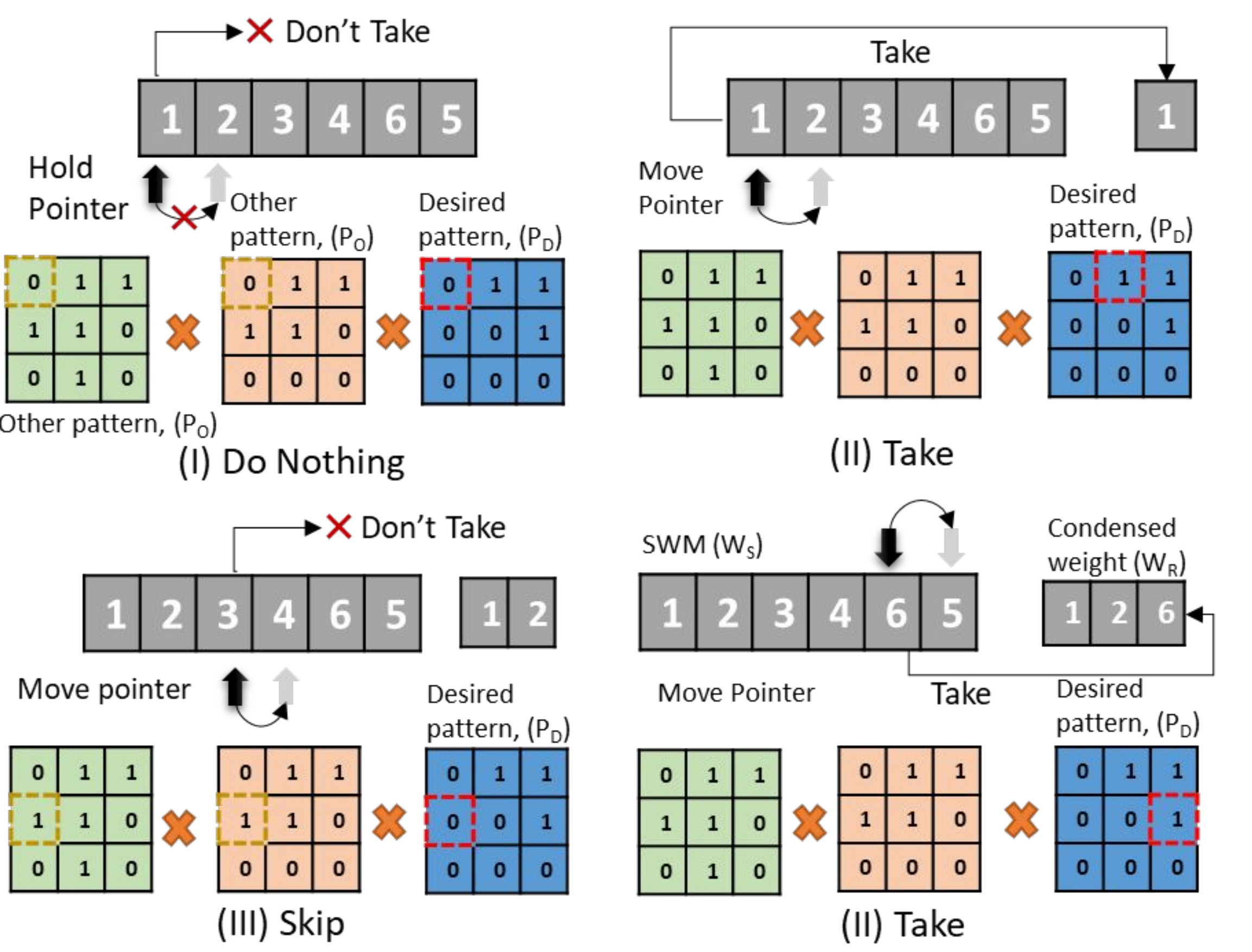}
\vspace{-15pt}
\caption{Weight extraction example \textcolor{black}{for low sparsity (blue) model}. \textcolor{black}{Based on the desired model pattern ($P_D$) and other model patterns ($P_O$), the desired model's weights are extracted with three basic steps including (I) Take, (II) Skip, and (III) Do Nothing, according to the patterns.}} 
\label{Fig:arch_extrac_more}
\end{figure}

\textcolor{black}{
To adapt to different environments, a particular model needs to be extracted from the \textit{SWM}. Therefore, we perform model reconstruction by extracting the model specific weights from the \textit{SWM}.
Algorithm \ref{alg:cap} and Figure \ref{Fig:arch_extrac_more} describes the model specific weight extraction process. The extracted weights are in condensed format ($W_{R}$) as shown in Figure \ref{Fig:arch_com_extra}. The extraction process has three elements including 1) \textit{SWM} ($W_{S}$), 2) Desired model pattern ($P_{D}$), and 3) Other shared model patterns ($P_{O}$).
Figure \ref{Fig:arch_extrac_more} shows that the algorithm includes three basic steps including 1) Take, 2) Skip, and 3) Do nothing. }

\textcolor{black}{
\textbf{Initial Setup}: The algorithm will operate those steps by iterating through the desired model pattern first. A source pointer points to the first element of the \textit{SWM}.}

\textcolor{black}{
\textbf{Take}: While iterating through the desired pattern $P_{D}$, a value of 1 means it will take the element that points to the \textit{SWM}. Next, the source pointer will move to the adjacent element. In Algorithm \ref{alg:cap}, line 4-7 perform this step.}

\textcolor{black}{
\textbf{Skip}: However, if the value of the desired pattern, $P_{D}$ is 0, there are two other cases. In the first case, the algorithm will investigate other shared-model patterns, $P_{O}$. If any of the other patterns has a value of 1 within same spot, then the source pointer will move to the next element but nothing will be taken from the \textit{SWM}. In Algorithm \ref{alg:cap}, line 9-12 perform this step.}

\textcolor{black}{
\textbf{Do Nothing}: In last case, the value of the desired pattern,  $P_{D}$ is 0 and all the other shared-model pattern, $P_{O}$ has a value of 0. Therefore, the source pointer will not be moved and nothing will be taken from the \textit{SWM}. }

\textcolor{black}{
Figure \ref{Fig:arch_extrac_more} shows an example where a high sparsity model is extracted from the \textit{SWM}. In this particular example, the \textit{SWM} is comprised of three different sparsity models. Similarly, low and medium sparsity models can also be extracted by selecting the low sparsity pattern and medium sparsity pattern as desired pattern ($P_{D}$) respectively. The Algorithm \ref{alg:cap} describes the whole process. 
Figure \ref{Fig:arch_com_extra} shows the extracted condensed weights that represents the individual sparsity model.}


\begin{algorithm}
\footnotesize
\caption{Weight Extraction Algorithm}\label{alg:cap}
\begin{algorithmic}[1]
\State $\textbf{Input: } W_{S},P_{D},[P_{O}]$
\State $\textbf{Output: } W_{R}$
\State $i, j \gets 0;$
\While{$p_{d} \in P_{D}$}
\If{$p_{d}$ is 1}
    \State $W_{R}[j] \gets W_{S}[i]$
    \State $i++;j++;$ 
\Else
    \While{$p_{o} \in [P_{O}]$ | $p_{o}$, $p_{d}$ with same index,}
        \If{$p_{o}$ is 1}
        \State $i++;$
        \State \textbf{break;}
        \EndIf
    \EndWhile
\EndIf
\EndWhile
\end{algorithmic}
\end{algorithm}

\subsubsection{Convolution}
\begin{figure}
\includegraphics[scale=0.33]{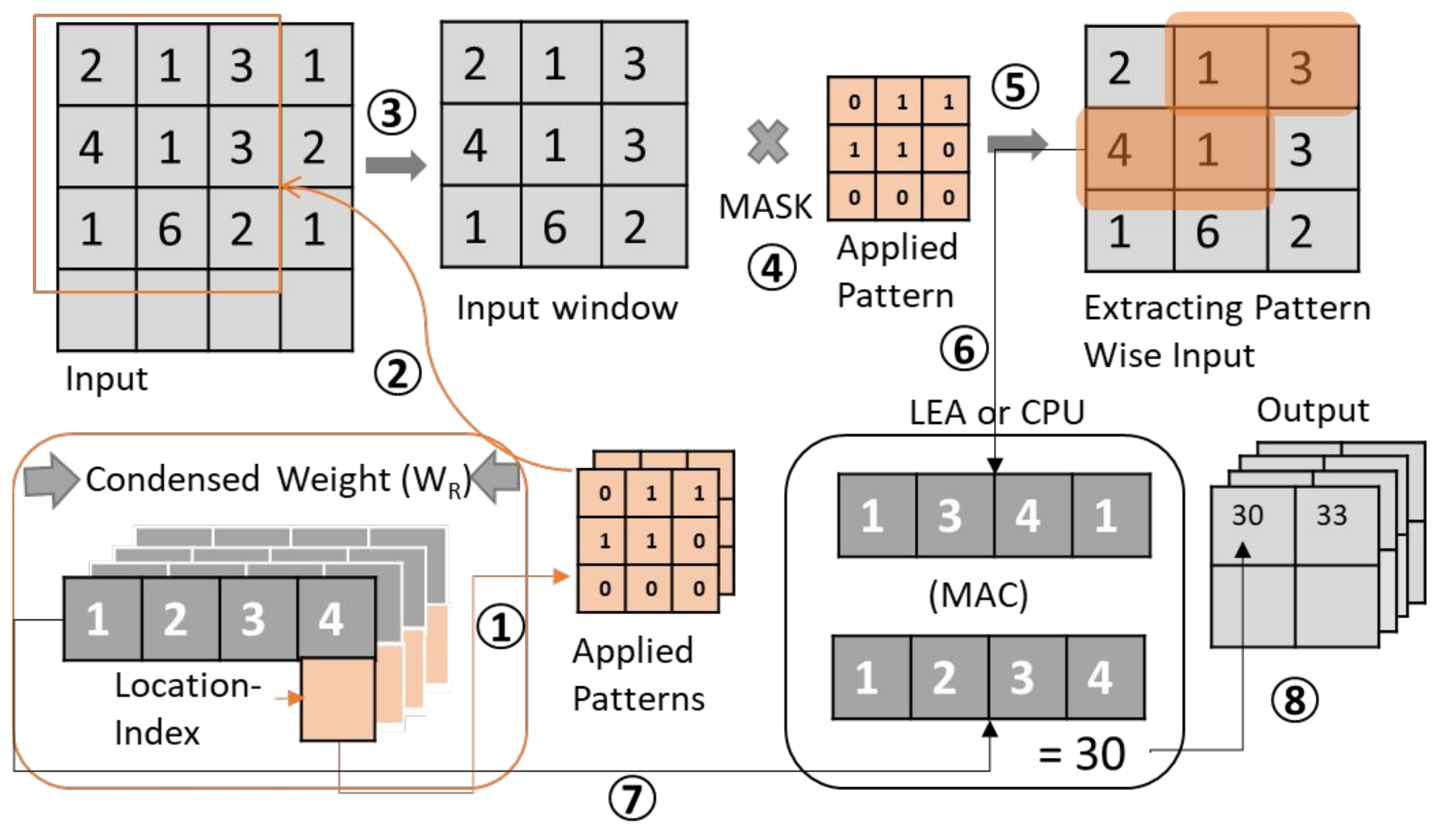}
\caption{Convolution with condensed weight ($W_R$):
based on the bit values of the pattern, corresponding input values are extracted and multiply-accumulation is performed to generate output.}
\label{Fig:arch_conv}
\end{figure}

\textcolor{black}{
We perform convolution operation with the extracted condensed weight during inference. Since the weights are condensed, we extract the corresponding input from the input window by applying the same pattern associated with the condensed weight as shown in Figure \ref{Fig:arch_conv}. Finally, the multiplication results are accumulated to the output.}

\subsubsection{Computation}
\textcolor{black}{
In on-device CPU based implementation, the multiply and accumulation operation takes place with single element by checking the pattern value. However, the on-device low energy accelerator (LEA) can expedite the computation by bulk operation. For a particular CNN computation window, the extracted condensed weight and the corresponding input values are transferred for computation. Since direct memory access (DMA) is a reliable and faster data transfer method, we use DMA to transfer these data from non-volatile memory (NVM) to volatile memory (VM). Finally, we invoke the MAC API from DSPLIB to perform the bulk computation. As the device is prone to frequent power failure, we implemented the inference with index based checkpointing  \cite{islamenabling} method so that the inference can resume when power restores.}

\subsubsection{Fully Connected Layer}
\textcolor{black}{
In fully connected (FC) layer the weight-sharing and extraction process is almost similar to the Convolutional layer except the fact that FC layer weights are usually large 2D matrix where convolutional weights can vary. However, the same shapes of patterns are applied to the large 2D weight. Thus, we divide the large 2D weight matrix into several blocks and performed similar process on these blocks as discussed above. For instance, in an FC layer if a weight matrix shape is m*n and the applied pattern shape is x*y, then there will be m/x*n/y number of blocks and every block of matrix will follow the similar compression, extraction and computation process.}



\subsection{Adaptive Inference}
\textcolor{black}{
Our framework is capable of switching \textit{instantly} between different sparsity models on run-time. The model switching is performed instinctively by understanding the environment and executing the model that is designed for that specific environment. To understand the environment, we designed $energyTraker$, a simple software solution to track energy availability. This is implemented with a timer that recurrently checks the voltage level of the device in parallel with the computation. A very low voltage level is recognized as a possible power outage, thus, calculating the current power cycle time. Similarly, we detect the last three power cycle times, and based on the pre-defined threshold value, we classify the current environment as high, medium, or low and adapt with the suitable model for the next inference.}

\vspace{-7pt}
\textcolor{black}{\subsection{Generality}
Although we demonstrated our work with three different sparsity models, our proposed framework can achieve an arbitrary number of shared-weights models. However, it is important to consider that increased shared-weight models will worsen the overhead since the escalating pattern information (Applied Pattern, Pattern-Index) will require more memory footprint and exacerbate the model extraction complexity.
Even though our experiment is conducted with convolutional neural network (CNN), this approach can be applied to other networks such as RNN, transformer, or any attention-based network. Since these networks consist of a large 2D weight matrix, similar process can be followed for the FC layer.}

\section{Experimental Evaluation}
\subsection{Hardware Setup}
The experiments are conducted with TI's MSP430FR5994 ultra-low-power evaluation board, consisting of a 16 MHz MCU, a 8KB volatile SRAM, a 256KB nonvolatile FRAM memory, and a low-energy Accelerator (LEA) that supports independent vector operations such as FFT, IFFT, MAC, etc.
The board is powered by an energy harvesting module composed of a function generator SIGLENT SDG1032X~\cite{sdg1032x}, a power regulator Bq25570, and an energy buffer (100µF capacitor). The function generator is used to simulate different energy harvesting power sources, which are insufficient for completing a single inference while causing frequent power failures. The power regulator provides a constant voltage of 3.3V for the normal operation of the board. TI EnergyTrace tool is used to measure energy consumption~\cite{EnergyTrace}. 

\vspace{3pt}
\noindent\textbf{DNN Models:}
This paper considers four DNN models as shown in Table~\ref{tab:re}. They are Image Classification (MNIST~\cite{lecun-mnisthandwrittendigit-2010} and ImageNet~\cite{imagenet}), Human Activity Recognition (HAR~\cite{anguita2013public_HAR}), and Google Keyword Recognition (OKG~\cite{Warden2018SpeechCA}) to represent image-based applications, wearable applications, and audio applications respectively.

\subsection{Experimental Results}

\subsubsection{Comparison of Shared Weight Training and AutoML Search:}

\begin{table*}[htbp]
  \centering
  \begin{footnotesize}
  \tabcolsep 5.2pt
  \renewcommand\arraystretch{1.1}
    \begin{tabular}{|cc|ccc|ccc|ccc|ccc|}
    \hline
    \multicolumn{2}{|c|}{Dataset/Task} & \multicolumn{3}{c|}{MNIST (Image Classification)} & \multicolumn{3}{c|}{HAR (Human Activity Recognition)} & \multicolumn{3}{c|}{OKG (Speech Recognition)} & \multicolumn{3}{c|}{ImageNet (Image Classification)} \\
    \hline
    \multicolumn{2}{|c|}{\multirow{2}[1]{*}{Models}} & \multicolumn{3}{c|}{LeNet-5~\cite{lecun1998gradient}} & \multicolumn{3}{c|}{HAR-Net~\cite{anguita2013public_HAR}} & \multicolumn{3}{c|}{OKG-Net~\cite{Warden2018SpeechCA}} & \multicolumn{3}{c|}{SqNxt-23~\cite{gholami2018squeezenext}}\\
    \multicolumn{2}{|c|}{} & M1 & M2 & M3 & M1 & M2 & M3 & M1 & M2 & M3 & M1 & M2 & M3 \\
    \hline
    \multirow{2}[1]{*}{SW-Train} & Sparsity & 60.24\% & 40.50\% & 20.70\% & 24.13\% & 5.31\% & 2.69\% & 33.33\% & 5.88\% & 5.02\% & 55.39\% & 54.30\% &54.07\%\\
    & Accuracy & 98.94\% & 98.96\% & 99.02\% & 87.21\% & 89.78\% & 90.19\% & 71.99\% & 73.82\% & 75.20\% & 73.80\% & 75.02\% &77.36\% \\
    & Latency (s) & 1.1 & 1.3 & 1.5 & 0.66 & 0.83 & 0.85 & 2.26 & 3.19 & 3.23 & 3.4 & 3.6 & 3.7\\
    \hline
    \multirow{2}[1]{*}{EVE} & Sparsity & 60.24\% & 48.00\% & 35.76\% & 24.19\% & 5.37\% & 2.69\% & 55.56\% & 25.85\% & 4.17\% & 57.15\% & 35.24\% &1.41\% \\
    & Accuracy & 99.04\% & 99.14\% & 99.11\% & 92.4\% & 92.39\% & 89.03\% & 72.34\% & 73.17\% & 74.73\% & 77.13\% & 77.87\% &79.34\% \\
    & Latency (s) & 1.1 & 1.23 & 1.45 & 0.65 & 0.81 & 0.85 & 1.51 & 2.59 & 3.31 & 3.4 & 3.5 &3.6\\
    \hline
    \multicolumn{2}{|c|} {\textbf{Accuracy gap}} & \textbf{0.1\%} & \textbf{0.98\%} & \textbf{0.09\%} & \textbf{2.21\%} & \textbf{2.61\%} & \textbf{1.82\%} & \textbf{0.52\%} & \textbf{0.24\%} & \textbf{0.6\%} &\textbf{3.33\%} & \textbf{2.85\%} &\textbf{1.98\%}\\
    \multicolumn{2}{|c|}{\textbf{Latency gap (s)}} & \textbf{0.00} & \textbf{0.07} & \textbf{0.05} & \textbf{0.01} & \textbf{0.02} & \textbf{0.00} & \textbf{0.01} & \textbf{0.00} & \textbf{0.00}
    &\textbf{0.00} & \textbf{0.10} &\textbf{0.10}\\
    \hline
    \end{tabular}%
    \end{footnotesize}
    \caption{Comparison between shared weight training and EVE}
  \label{tab:re}%
  \vspace{-10pt}
\end{table*}%

\begin{figure}
     \begin{subfigure}[b]{0.45\columnwidth}
         \includegraphics[width=4.2cm, height=3.5cm]{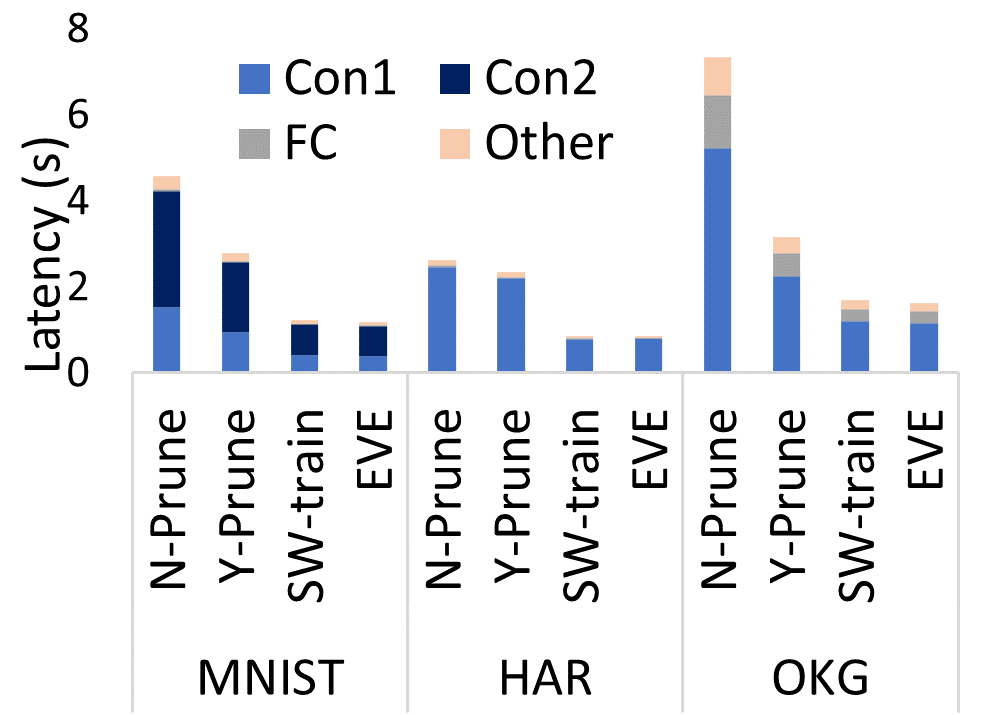}
          \vspace{-10pt}
          \hspace{-10pt}
         \caption{}
         \label{fig:}
     \end{subfigure}
     \hfill
     \begin{subfigure}[b]{0.45\columnwidth}
        \vspace{-10pt}
         \hspace{-20pt}
         \includegraphics[width=4.2cm, height=3.5cm]{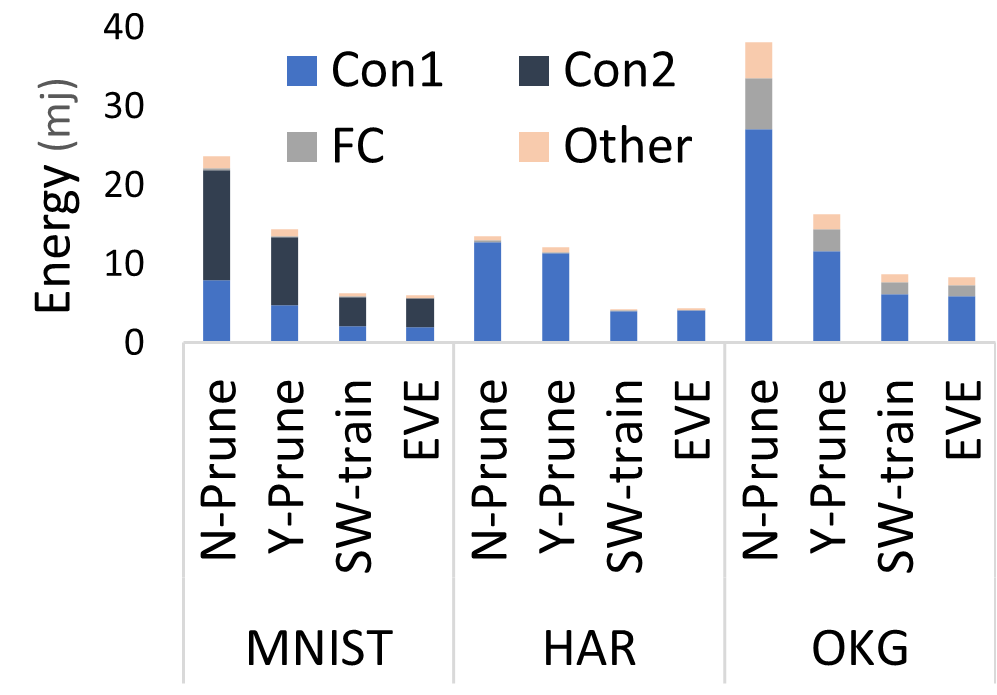}
         \label{fig:}
         \caption{}
     \end{subfigure}
     \vspace{-10pt}
        \caption{(a) Latency and (b) Energy under continuous power.}
        \label{Fig:latency_cont_power}
\end{figure}


\begin{figure}
     \begin{subfigure}[b]{0.32\columnwidth}
         \includegraphics[width=2.7cm, height=3.0cm]{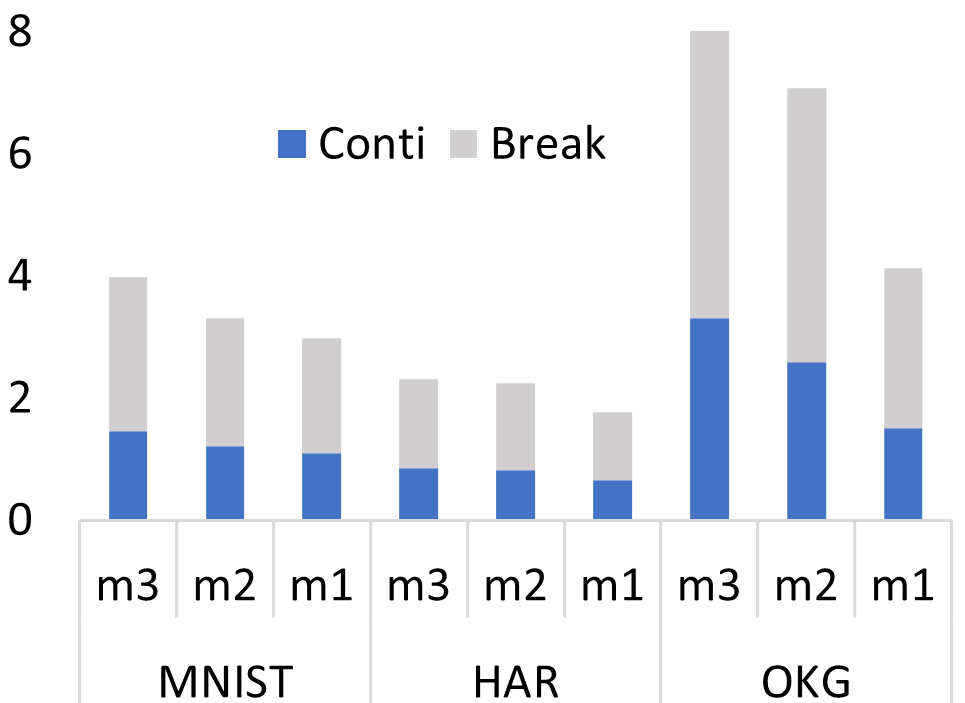}
         \caption{}
         \label{fig:}
     \end{subfigure}
     \hfill
     \begin{subfigure}[b]{0.32\columnwidth}
         \hspace{-10pt}
         \includegraphics[width=2.5cm, height=3.0cm]{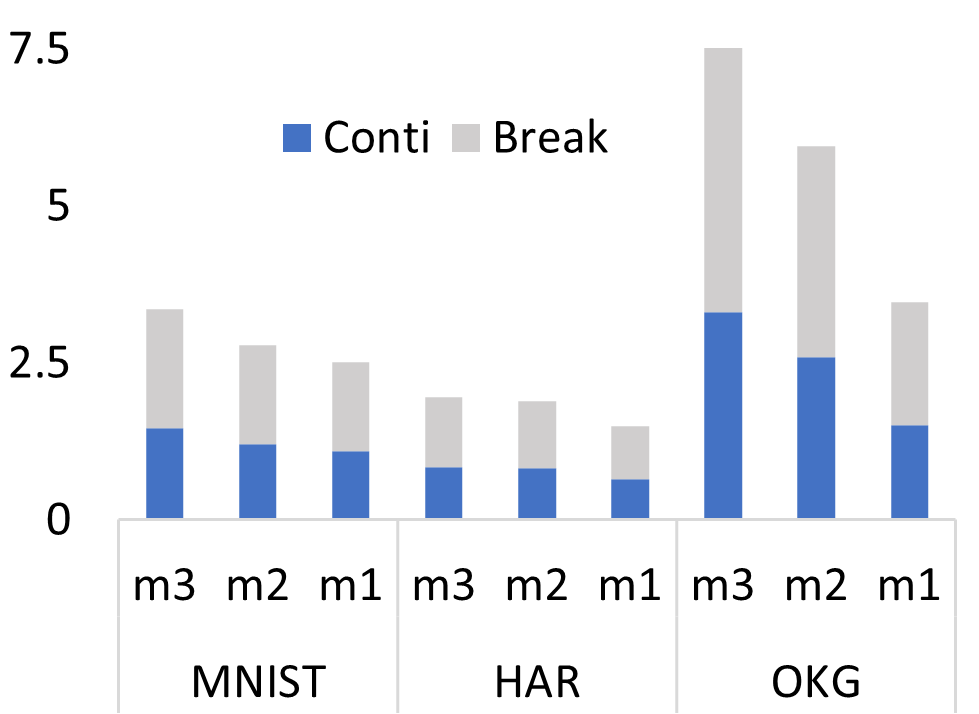}
         \label{fig:}
         \caption{}
     \end{subfigure}
     \hfill
     \begin{subfigure}[b]{0.32\columnwidth}
         \hspace{-20pt}
         \includegraphics[width=2.7cm,, height=3.0cm]{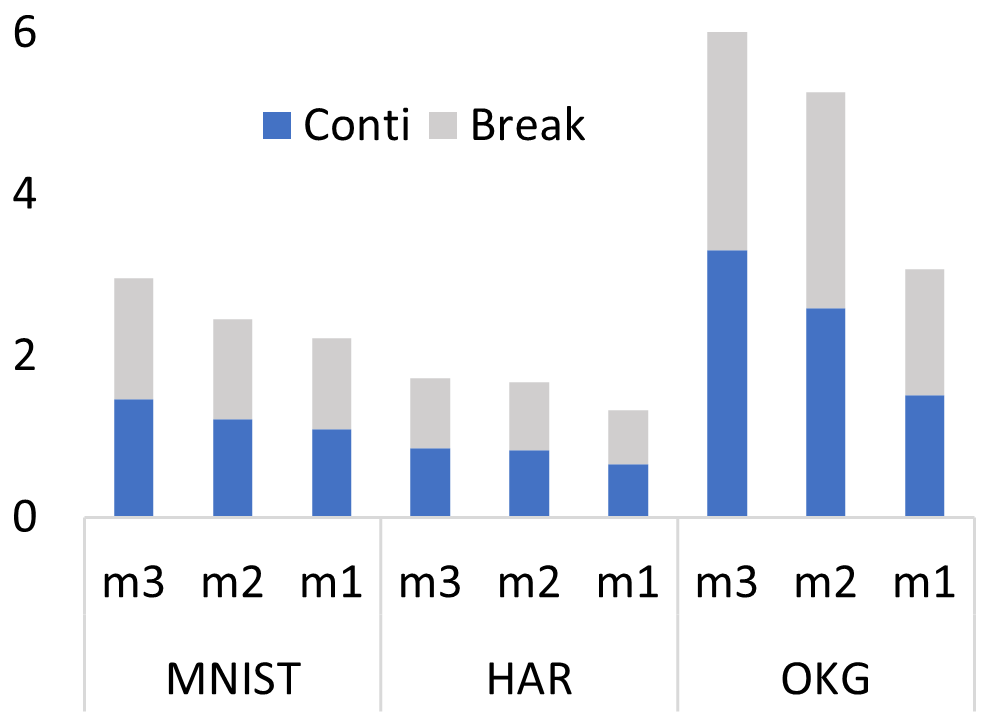}
         \label{fig:}
         \caption{}
     \end{subfigure}
     \vspace{-10pt}
         \caption{(a) Latency under 3mW  (b) Latency under 4mW. (c) Latency under 5mW.}\label{Fig:latency_different_power}
\end{figure}




In Table~\ref{tab:re}, the row \textit{SW-Train} shows the accuracy of different sparsities after shared-weights training. We can tell from the table that shared-weights training can help retain weights among different sparsities to have them shared with a comparable accuracy. The patterns applied on the weight matrix are selected manually from the pattern searching space according to the rule proposed in the PCONV paper\cite{Ma2020PCONVTM}. 
As for LeNet5, the high sparsity is 60.24\% and the corresponding accuracy is 98.94\%. The medium Sparsity is 20.093\% and its accuracy is 98.96\%. The low sparsity is 8.47\% and its accuracy is 99.02\%. 
In order to ensure ``shared weight" as illustrated in Figure~\ref{Seq}, we try to ensure that there is no overlap of rows or columns between patterns for models under different sparsity levels when manually selecting those patterns. 
Apart from image classification tasks, our shared-weights training shows satisfying accuracy on recognition tasks as well. For example, as for the HAR-Net model, the high, medium and low sparsities are 24.13\%, 5.31\% and 2.69\%, respectively, and their corresponding accuracy are 90.19\%, 89.78\% and 87.21\%. Comparably, the accuracy and sparsity of the OKG-Net show the same trend as HAR-Net. Thus, with sacrificing little accuracy drop, the shared-weights training algorithm could keep weights unchanged, and further benefit the inference process on energy harvesting devices. 

The second row (EVE) in Table~\ref{tab:re} shows the accuracy and model latency under three different sparsities after shared-weights training, with the three pruning patterns are automatically selected by the AutoML search algorithm. For LeNet-5 and OKG-Net, as there are totally 44 patterns in the pattern space, the total dimension of the search space for the three models is $ 44 \times 44 \times 44=85184$. For HAR-Net, as there are totally 21 patterns in the pattern space, the total dimension of the search space for the three models is  $21 \times 21 \times 21=9261$. For SqNxt-23, as there are 36 patterns in the pattern space, the total dimension and the total of the search space is $36 \times 36 \times 36=46659$. 

These action spaces are too large for human selection. That is why we use a Reinforcement Learning algorithm to design the AutoML search. Our RNN-based agent has one layer with $35$ hidden units and a fully-connected layer with dimension-width size, then finally after a softmax function. The agent can choose the corresponding pattern with the highest probability. The batch size for updating the RNN-agent is $1$. Thus, after the agent gives an output action, the RNN agent would be updated by the policy gradient algorithm. When the models with selected three patterns satisfy the constraints of both accuracy and latency, the searching algorithm will return the models and the selected three pruning patterns. If not, the agent will continue exploring until $300$ episodes. 

\subsubsection{Inference Time under Continuous Power Supply:}
To fairly evaluate the performance of the proposed methods, we demonstrate the performance of having 3 models with shared weights on the device. \textcolor{black}{Since to the best of our knowledge, shared weight training is a new idea that has not been implemented before on energy harvesting (EH) device,} we first compared the inference latency of executing 3 generated models with shared weights of the two proposed methods (SW-train and EVE) in a sequence with two baselines for each benchmark. Baseline 1 is named as N-prune, where each backbone model without pruning is stored on the off-chip memory and has to be loaded into the on-chip memory for execution. Baseline 2 is named Y-prune, where we can only fit one pruned backbone model into the on-chip memory. We can observe from the Figure~\ref{Fig:latency_cont_power} that without using any pruning method, executing a DNN backbone model costs extremely long latency of 4s, 2.5s ,8s respectively. Y-prune generates three pruned models but cannot fit them all into the on-chip memory, and thus require less time than SW-train but more time than SW-train and EVE. We can observe from the Figure~\ref{Fig:latency_cont_power}(a) that SW-train and EVE run $2.5\times$ and $2.2\times$ faster than the two baselines.

\subsubsection{Energy Consumption under Continuous Power Supply:}
In terms of energy consumption evaluation, SW-train and EVE also outperform the two baselines. As shown in Figure~\ref{Fig:latency_cont_power} (b) that, SW-train and EVE achieve $2.5\times$ and $2.77\times$ energy-saving on MNIST; $2.8\times$ and $3.1\times$ energy-saving on HAR. And finally, $3.2\times$ and $3.3\times$ energy is saved on OKG. Having shared models significantly reduced the energy for switching models from the off-chip memory.


\subsubsection{Inference Time under Intermittent Power Supply:}
To evaluate the performance of 3 models with shared weights generated with the proposed SW-Train and EVE methods, we use the function generator SIGLENT SDG1032X to generate different levels of harvesting power including \textcolor{black}{ 5mW, 4mW, 3mW as the high, medium, and low energy levels}. The three harvesting power are smaller than the working power of the energy harvesting device. Thus the device has to accumulate energy first before starting each working cycle of inference. Figure~\ref{Fig:latency_different_power} shows the inference latency when the harvesting is 5mW, 4mW, 3mW respectively for each of the three models with shared weights. From the three figures, we observe that, when the harvesting power is low, the execution time is significantly increased. Therefore, in order to meet the QoS requirement, the energy harvesting device needs to automatically switch to the low-accuracy model which takes much less time.




\subsubsection{Evaluation of Weight Reconstruction and Memory Overhead}
The weight reconstruction from the shared three models stored in on-chip memory brings 1 to 3\% overhead to the overall inference for each benchmark which is negligible considering the significantly improved overall performance.

\section{Conclusion}

In this work, we propose EVE, a novel pattern pruning based framework that generates multiple hardware friendly models with shared weights for energy harvesting devices. We develop an 
{\textcolor{black}{AutoML-based}}
co-exploration framework to search the desired multi models with shared weights while satisfying both the accuracy and latency constraints.
The generated models can successfully fit within the on-chip memory budget.
We develop an efficient on-device implementation architecture that compresses multiple shared kernels and extracts the kernels for corresponding individual model according to energy levels. 
Experimental results show that our design achieves $2.5\times$ speedup than the baseline. EVE can further achieve 1.2\% higher accuracy with higher sparsity compared to human based shared weight search.
It is worth noting that, the proposed EVE framework is generalized and can generate more than three models with shared weights to adapt to different energy harvesting power levels.

\section*{Acknowledgement}
The authors would like to acknowledge the USDOT Transportation Consortium of the South-Central States (TRAN-SET) (\# 21-034 to YFJ and MX, and \#21-049 to YFJ and MX), National Science Foundation (EEC-2051113 to YFJ, CCF-2011236 to WW, and CCF-2006748 to WW), USDA-NIFA Agriculture and Food Research Initiative (Award No.: 2022-67023-36399 to CD) for the funding and necessary support in completing the research. The funding sources had no role in the design of the study; collection, analysis, and interpretation of data; or in writing the manuscript.

\newpage
\bibliographystyle{ACM-Reference-Format}
\bibliography{references,ref}

\end{document}